\documentclass[runningheads]{llncs}
\usepackage{multicol}
\usepackage[T1]{fontenc}
\usepackage{graphicx}
\usepackage{bm}
\usepackage{hyperref}
\usepackage{xcolor}
\usepackage{url} 

\begin{document}
\title{GANeXt: A Fully ConvNeXt-Enhanced Generative Adversarial Network for MRI- and CBCT-to-CT Synthesis}
\titlerunning{GANeXt for MRI- and CBCT-to-CT Synthesis}
%
\author{
Siyuan Mei\inst{1}\textsuperscript{*} \and
Yan Xia\inst{2}\textsuperscript{*} \and
Fuxin Fan\inst{3} \and
Andreas Maier\inst{1}
}
\institute{
Pattern Recognition Lab, Friedrich-Alexander-University Erlangen-Nuremberg, Erlangen, Germany \and
Department of Orthodontics and Orofacial Orthopedics, Friedrich-Alexander-University Erlangen-Nuremberg, Erlangen, Germany \and
Digital Technology and Innovation, Siemens Healthineers, Shanghai, China \\
\textsuperscript{*}These authors contributed equally to this work.
}

\authorrunning{S. Mei et al.}
%
%
\maketitle              
\begin{abstract}
The synthesis of computed tomography (CT) from magnetic resonance imaging (MRI) and cone-beam CT (CBCT) plays a critical role in clinical treatment planning by enabling accurate anatomical representation in adaptive radiotherapy. In this work, we propose GANeXt, a 3D patch-based, fully ConvNeXt-powered generative adversarial network for unified CT synthesis across different modalities and anatomical regions. Specifically, GANeXt employs an efficient U-shaped generator constructed from stacked 3D ConvNeXt blocks with compact convolution kernels, while the discriminator adopts a conditional PatchGAN. To improve synthesis quality, we incorporate a combination of loss functions, including mean absolute error (MAE), perceptual loss, segmentation-based masked MAE, and adversarial loss and a combination of Dice loss and cross-entropy for multi-head segmentation discriminator. For both tasks, training is performed with a batch size of 8 using two separate AdamW optimizers for the generator and discriminator, each equipped with a warmup and cosine decay scheduler, with learning rates of $5\times10^{-4}$ and $1\times10^{-3}$, respectively. Data preprocessing includes deformable registration, foreground cropping, percentile normalization for the input modality, and linear normalization of the CT to the range $[-1024, 1000]$. Data augmentation involves random zooming within $(0.8, 1.3)$ (for MRI-to-CT only), fixed-size cropping to $32\times160\times192$ for MRI-to-CT and $32\times128\times128$ for CBCT-to-CT, and random flipping. During inference, we apply a sliding-window approach with $0.8$ overlap and average folding to reconstruct the full-size sCT, followed by inversion of the CT normalization. After joint training on all regions without any fine-tuning, the final models are selected at the end of 3000 epochs for MRI-to-CT and 1000 epochs for CBCT-to-CT using the full training dataset.

\end{abstract}

\section{Introduction}

The generation of synthetic computed tomography (sCT) from magnetic resonance imaging (MRI) or cone-beam CT (CBCT) has become an essential task in modern radiotherapy workflows, particularly for adaptive treatment planning where accurate representation of patient anatomy is crucial~\cite{report,thummerer2023synthrad2023}. In recent years, a number of attention-based architectures, such as SwinUNETR~\cite{hatamizadeh2021swin} and its variants, have been explored for sCT synthesis~\cite{report}, leveraging long-range dependency modeling to capture global anatomical context. While these approaches have demonstrated competitive results, emerging evidence from large-scale 3D medical image analysis suggests that purely convolutional architectures remain highly effective, and in certain cases superior, for volumetric data processing~\cite{isensee2024nnure,roy2023mednext}. Notably, MedNeXt~\cite{roy2023mednext} has shown state-of-the-art performance in 3D medical image segmentation, highlighting the potential of modernized convolutional designs~\cite{woo2023convnext} to outperform transformer-based models in terms of accuracy, efficiency, and robustness.

Motivated by these findings, we design GANeXt, a fully convolutional generative adversarial network tailored for unified CT synthesis across modalities and anatomical regions. At its core, GANeXt employs GeNeXt, an improved MedNeXt~\cite{roy2023mednext} generator constructed entirely from modern 3D ConvNeXt~\cite{woo2023convnext} blocks, replacing conventional U-Net~\cite{ronneberger2015unet} convolutions with enhanced depthwise convolution and feed-forward modules that improve representation power while maintaining computational efficiency. This modernized convolutional backbone, coupled with a conditional PatchGAN discriminator~\cite{pix2pix} and a carefully designed multi-loss training objective, enables GANeXt to achieve substantial performance gains in sCT generation without relying on attention mechanisms.

\begin{figure}[tb]
\centering
\includegraphics[width=0.9\textwidth]{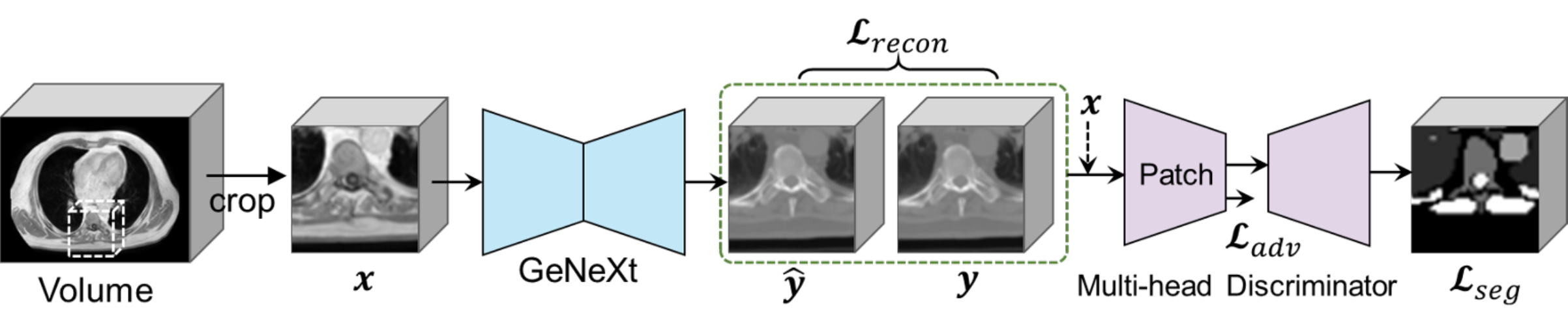}
\caption{Overview of GANeXt} \label{ganext}
\end{figure}

\section{Methods}

\begin{figure}[htb]
\centering
\includegraphics[width=\textwidth]{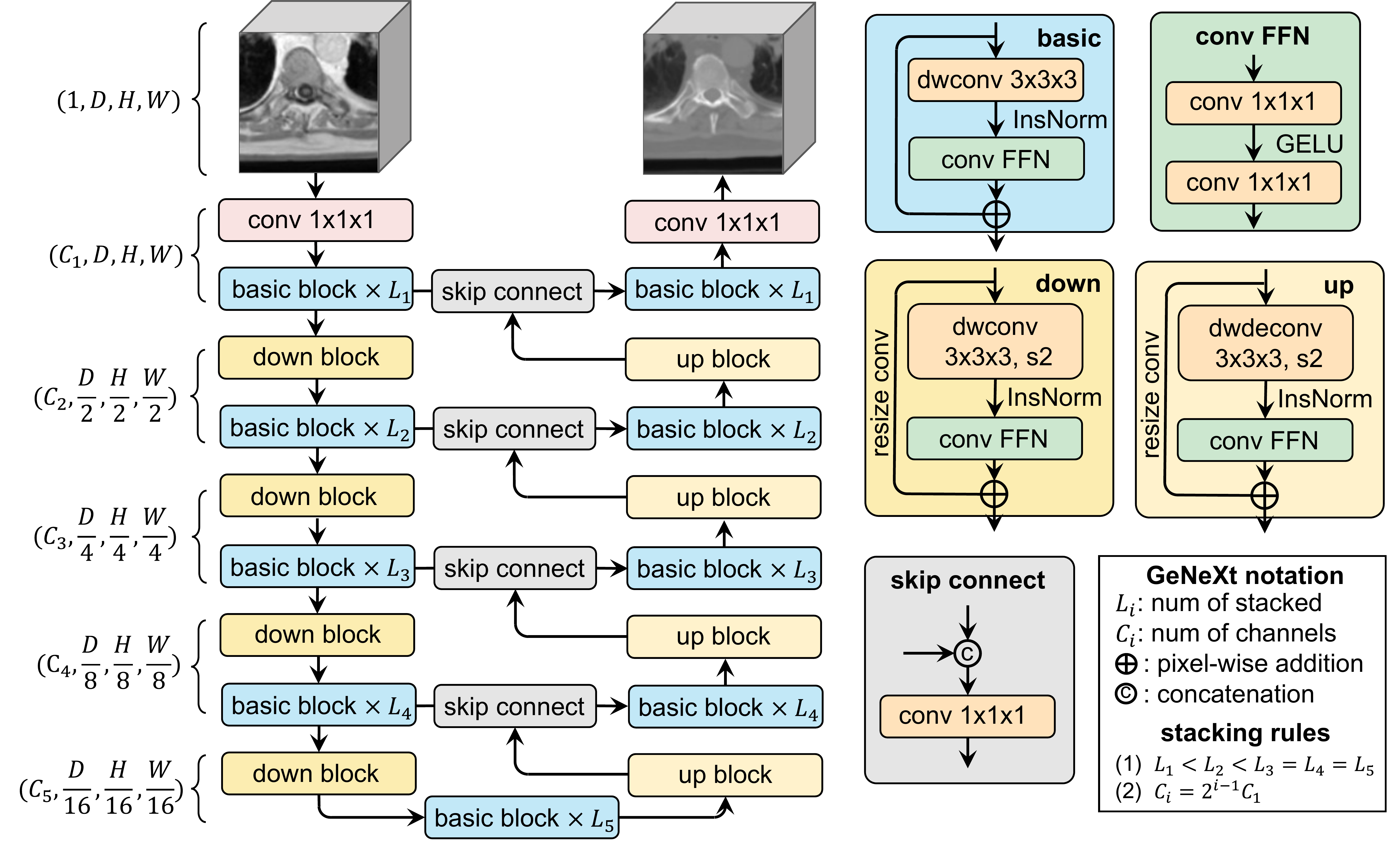}
\caption{Structure of GeNeXt, which integrates modern modifications of ConvNet into the 3D U-Net.} \label{genext}
\end{figure}

\subsection{Data and preprocessing}

This study utilizes the dataset from the SynthRAD2025 Grand Challenge ~\cite{thummerer2025synthrad2025}, a benchmark platform designed to compare synthetic CT (sCT) generation methods using public data and standardized evaluation metrics. The challenge is structured around two key tasks: MRI-to-sCT generation, aimed at facilitating MRI-only and MRI-based adaptive radiotherapy, and CBCT-to-sCT generation to support CBCT-based adaptive radiotherapy. The dataset comprises 2362 cases, where 890 are MRI-CT pairs for task 1 and 1472 are CBCT-CT pairs for task 2.

First, we randomly split 15\% of the training data as the validation dataset. Then all CT data undergo deformable registration to match the challenge test set. For both tasks, preprossessing involves percentile normalization of the input modality using the 99.5\% threshold, linear normalization of CT intensities using the min/max value of -1024/1000 for MR to CT image generation and the min/max value of -1024/1500 for for CBCT to CT generation, and foreground cropping based on the body mask. 

\subsection{Model}

As illustrated in Figure~\ref{ganext}, the input volume is first cropped to a 3D patch, which is then fed into a GeNeXt generator to generate a synthetic CT patch. Next, a conditional PatchGAN discriminator~\cite{pix2pix} is used to distinguish the synthetic CT and the real CT patches. The whole model is implemented with PyTorch and the MONAI framework~\cite{cardoso2022monai}. We describe the detailed structures of the generator and discriminator below.

\subsubsection{GeNeXt Generator}

Figure~\ref{genext} depicts the U-shaped GeNeXt architecture, which follows an encoder–decoder paradigm with five hierarchical stages. Given a single-channel input patch of shape $1\times D\times H\times W$, the first encoder stage maps it to $C_1$ channel using a 3D $1\times 1\times 1$ convolution layer, followed by a stack of GeNeXt basic blocks. Subsequent encoder stages progressively downsample the spatial resolution by a factor of $2$ and double the channel capacity using downsampling blocks and basic blocks. In the symmetric decoder, the feature maps are progressively restored to higher resolutions via upsampling blocks and basic blocks, while skip connections fuse encoder and decoder features at corresponding resolutions. A final $1\times 1\times 1$ convolution maps the output channels from $C_1$ back to $1$. The key building blocks are as follows:

\begin{enumerate}
    \item GeNeXt basic block: The block begins with a depthwise convolution layer. Unlike the large kernel size used in 2D ConvNeXt, we adopt a $3\times 3\times 3$ kernel for 3D medical images and replace the subsequent layer normalization with instance normalization to better accommodate small batch sizes. This is followed by a transformer-inspired convolutional feed-forward network (convFFN) that enhances representational capacity, where the first pointwise $1\times 1\times 1$ convolution expands the channels and the second $1\times 1\times 1$ convolution projects them back to the original dimensionality. The expansion ratio is stage-specific and denoted as $R_i$. A residual connection is employed by adding the block's input to the conv FFN output, facilitating stable gradient flow and residual learning.
    \item GeNeXt downsampling block: This block modifies the first layer of the basic block into a stride-$2$ depthwise convolution for spatial downsampling. To preserve residual learning, the shortcut path incorporates a resize convolution layer~\cite{resizeconv} that matches the spatial and channel dimensions before addition.  
    \item GeNeXt upsampling block: Analogous to the downsampling block, the upsampling block replaces the first layer with a stride-$2$ transposed convolution for spatial upsampling, while retaining the remaining layers of the basic block.
    \item Skip connection block: To fuse encoder and decoder features, we concatenate the encoder output and the upsampled decoder feature map at the same stage, followed by a $1\times 1\times 1$ convolution to project the concatenated features back to the target channel dimension.
\end{enumerate}

For both tasks, we configure the $C_1$ as 32, and the stage depth $\{L_1, L_2, L_3, L_4, L_5\}$ and the expansion ratio $\{R_1, R_2, R_3, R_4, R_5\}$ are set as $\{3,4,6,6,6\}$. In this configuration, the generator contains a total of \textbf{36.48M} parameters.

\subsubsection{Conditional PatchGAN Discriminator} 

The discriminator employs a 3D conditional PatchGAN~\cite{pix2pix} design, which takes the input patch $\bm{x}$ as an additional condition. Specifically, it comprises a cascade of strided $4\times4\times4$ convolutional layers with LeakyReLU activations and instance normalization, followed by a final $1\times1\times1$ convolution to generate patch-level real/fake scores of shape $1\times \frac{D}{16} \times \frac{H}{16} \times \frac{W}{16}$. The network progressively expands the channel capacity from $32$ to $512$ while downsampling the spatial resolution, resulting in a final receptive field that covers a local volumetric region rather than the entire image. This patch-based formulation enables the discriminator to emphasize high-frequency structures and local texture realism in 3D volumes. The total number of parameters is \textbf{11.15M}.

\subsubsection{Multi-Head SegPatchGAN Discriminator} 

For the CBCT to CT image generation task, we also employ an encoder-decoder discriminator with two heads: adversarial head and segmentation head, to encourage the generator to produce images that are more realistic and structurally coherent. Specifically, we use the existing downsampling path of the PatchGAN discriminator to extract features at different scales and build a new upsampling path (a decoder) that takes features from the encoder and reconstructs a segmentation map. Both real CT images and generated CT images are used for training. When real images are fed into the discriminator, the discriminator produces a segmentation map, which is compared to the ground-truth segmentation map. The gradients from a Dice loss $\mathcal{L}_{\mathrm{seg}_d}$  are used to update the discriminator's weights. The discriminator's weights are frozen when generated, when fake CT images are fed into the discriminator. The gradients from $\mathcal{L}_{\mathrm{seg}_g}$ propagate backward through the frozen discriminator and are used to update the generator's weights, forcing the generator to learn semantic and structural information.

\subsection{Loss Function}

We employ a set of complementary loss functions to optimize the fidelity and perceptual quality of the synthetic CT volumes:

\begin{itemize}
\item \textbf{Mean Absolute Error (MAE) loss} $\mathcal{L}_{\mathrm{mae}}$: voxel-wise $\ell_1$ loss between the synthetic and target CT volumes, which enforces overall intensity accuracy.
\item \textbf{Perceptual loss} $\mathcal{L}_{\mathrm{perc}}$: a learned perceptual image patch similarity loss, where the pretrained VGG network~\cite{vgg} in the standard LPIPS~\cite{lpips} is replaced by a more powerful ConvNeXt-B~\cite{woo2023convnext} backbone to better capture structural and texture information.
\item \textbf{Segmentation-based masked MAE loss} $\mathcal{L}_{\mathrm{mask}}$: MAE computed only within anatomical regions segmented by a pretrained TotalSegmentator model~\cite{wasserthal2023totalsegmentator}, focusing supervision on clinically relevant structures.
\item \textbf{Adversarial loss} $\mathcal{L}_{\mathrm{adv}}$: a conditional GAN loss that encourages the generator to produce outputs indistinguishable from real CT volumes when evaluated by the PatchGAN discriminator.
\item \textbf{Feature matching loss} $\mathcal{L}_{\mathrm{fm}}$: applied to the discriminator’s intermediate activations to stabilize adversarial training and promote convergence by aligning high-level feature representations between real and synthetic CTs.
\item \textbf{Segmentation D loss} \(\mathcal{L}_{\mathrm{seg}_d}\): a predicted segmentation map reconstructed from real image is compared against the ground truth segmentation map using a combination of Dice loss and cross-entropy to train the segmentation head of the discriminator to correctly segment CT images.
\item \textbf{Segmentation G loss} \(\mathcal{L}_{\mathrm{seg}_g}\): a predicted segmentation map from generated CT image is compared against the ground truth segmentation map using the combined segmentation loss to update the generator’s weights, to encourage the generator to produce images that are both realistic and structurally consistent.

\end{itemize}

For MRI to CT generation task, the total objective is a weighted sum of these components : 

\begin{equation}
\mathcal{L}_{\mathrm{total}} 
= \lambda_{1}\mathcal{L}_{\mathrm{mae}} 
+ \lambda_{2}\mathcal{L}_{\mathrm{perc}} 
+ \lambda_{3}\mathcal{L}_{\mathrm{mask}} 
+ \lambda_{4}\mathcal{L}_{\mathrm{adv}} 
+ \lambda_{5}\mathcal{L}_{\mathrm{fm}} ,
\end{equation}
where $\lambda_{1}:\lambda_{2}:\lambda_{3}:\lambda_{4}:\lambda_{5}=10:1:50:10:10$.

For CBCT to CT generation task, the total objective is a weighted sum of these components : 
\begin{equation}
\mathcal{L}_{\mathrm{total}} 
= \lambda_{1}\mathcal{L}_{\mathrm{mae}} 
+ \lambda_{2}\mathcal{L}_{\mathrm{perc}} 
+ \lambda_{3}\mathcal{L}_{\mathrm{mask}} 
+ \lambda_{4}\mathcal{L}_{\mathrm{adv}} 
+ \lambda_{5}\mathcal{L}_{\mathrm{fm}}
+ \lambda_{6}\mathcal{L}_{\mathrm{seg}_d}\
+ \lambda_{7}\mathcal{L}_{\mathrm{seg}_g}\,
\end{equation}
where $\lambda_{1}:\lambda_{2}:\lambda_{3}:\lambda_{4}:\lambda_{5}:\lambda_{6}:\lambda_{7}=10:1:10:10:10:0.5:0.5$.

We do not perform extensive fine-tuning of the weighting coefficients $\lambda_{\ast}$; instead, they are chosen empirically so that the magnitudes of the individual loss terms are approximately balanced during training.

\subsection{Training}

We adopt an identical training strategy for both tasks. The augmentation pipeline consists of random zooming within the range $(0.8, 1.3)$ (for task 1 only), random cropping to a fixed patch size of $32\times 160\times 192$ for task 1 and  $32\times 128\times 128$ for task 2, and random horizontal and vertical flipping. Training is conducted with a total batch size of 8 distributed across four A100 GPUs. For task 1, we employ two separate AdamW optimizers for the generator and discriminator, each coupled with a warmup and cosine decay learning rate scheduler. For task 2, we employ two separate Adam optimizers for the generator and discriminator. The initial learning rates for both tasks are set to $5\times 10^{-4}$ for the generator and $1\times 10^{-3}$ for the discriminator. Models are trained jointly on all anatomical regions without region-specific fine-tuning, and the final checkpoints are selected at the end of 3000 epochs for MRI-to-CT and 1000 epochs for CBCT-to-CT. 

\subsection{Inference}

At inference, we generate full-size sCT volumes using a sliding-window approach with an overlap ratio of $0.8$ and average folding for seamless reconstruction. The resulting sCT is then invert-normalized to restore the original HU values.

\subsection{Evaluation}

Model performance is quantitatively evaluated on the body regions using three masked image quality metrics: masked mean absolute error (MAE), masked multi-scale structural similarity index measure (MS-SSIM), and masked peak signal-to-noise ratio (PSNR).

\section{Results}

\begin{table}[htb]
\centering
\setlength{\tabcolsep}{12pt}
\renewcommand{\arraystretch}{1.15} 
\caption{Ablation study of different loss components on the MRI-to-CT validation set.}
\begin{tabular}{lccc}
\hline
\textbf{Loss} & \textbf{MAE} $\downarrow$ & \textbf{PSNR} $\uparrow$ & \textbf{MS-SSIM} $\uparrow$ \\
\hline
$\mathcal{L}_{\mathrm{mae}}$            & 74.8621$\pm$17.1052 & 27.1024$\pm$2.3128 & 0.9073$\pm$0.0118 \\
$+\mathcal{L}_{\mathrm{adv}}$           & 70.2435$\pm$18.4567 & 28.2148$\pm$2.4015 & 0.9168$\pm$0.0124 \\
$+\mathcal{L}_{\mathrm{fm}}$            & 69.4827$\pm$19.3054 & 28.4125$\pm$2.4562 & 0.9196$\pm$0.0129 \\
$+\mathcal{L}_{\mathrm{perc}}$          & 68.1243$\pm$20.2841 & 28.7684$\pm$2.4897 & 0.9229$\pm$0.0135 \\
$+\mathcal{L}_{\mathrm{mask}}$          & \textbf{67.4066$\pm$22.1124} & \textbf{29.0467$\pm$2.5285} & \textbf{0.9245$\pm$0.0139} \\
\hline
\end{tabular}
\label{tab:loss_ablation}
\end{table}

\begin{table}[htb]
\centering
\setlength{\tabcolsep}{5pt}
\renewcommand{\arraystretch}{1.15}
\caption{Comparison of different generator architectures on MRI-to-CT and CBCT-to-CT validation sets. Official challenge validation results are marked with $^\dagger$.}
\begin{tabular}{lccc}
\hline
\textbf{Model} & \textbf{MAE} $\downarrow$ & \textbf{PSNR} $\uparrow$ & \textbf{MS-SSIM} $\uparrow$ \\
\hline
\multicolumn{4}{l}{\textit{MRI-to-CT}} \\
3D U-Net~\cite{ronneberger2015unet}        & 78.4213$\pm$16.8542 & 26.4821$\pm$2.2145 & 0.9062$\pm$0.0113 \\
SwinUNETR~\cite{hatamizadeh2021swin}       & 76.3524$\pm$17.4215 & 26.9184$\pm$2.2514 & 0.9094$\pm$0.0118 \\
3D U-Net++~\cite{unetpp++}      & 74.2841$\pm$18.0543 & 27.3415$\pm$2.3214 & 0.9142$\pm$0.0121 \\
MedNeXt~\cite{roy2023mednext}         & 69.4523$\pm$21.3412 & 28.2325$\pm$2.4621 & 0.9201$\pm$0.0137 \\
\textbf{GANeXt (ours)} & \textbf{67.4066$\pm$22.1124} & \textbf{29.0467$\pm$2.5285} & \textbf{0.9245$\pm$0.0139} \\
\textbf{GANeXt$^\dagger$ (ours)} & \textbf{62.6230$\pm$22.7100} & \textbf{29.9792$\pm$2.5405} & \textbf{0.9325$\pm$0.0646} \\
\hline
\multicolumn{4}{l}{\textit{CBCT-to-CT}} \\
3D U-Net~\cite{ronneberger2015unet}        & 83.1542$\pm$15.9342 & 25.8421$\pm$2.1043 & 0.9024$\pm$0.0109 \\
SwinUNETR~\cite{hatamizadeh2021swin}       & 81.2045$\pm$16.4021 & 26.3145$\pm$2.1482 & 0.9067$\pm$0.0114 \\
3D U-Net++~\cite{unetpp++}      & 79.1123$\pm$17.0421 & 26.7542$\pm$2.2045 & 0.9112$\pm$0.0119 \\
MedNeXt~\cite{roy2023mednext}         & 73.8421$\pm$19.8542 & 27.9865$\pm$2.3562 & 0.9234$\pm$0.0131 \\
\textbf{GANeXt (ours)} & \textbf{52.1145$\pm$10.319} & \textbf{31.2487$\pm$1.7241} & \textbf{ 0.96621$\pm$0.0167} \\
\textbf{GANeXt$^\dagger$ (ours)} & \textbf{48.9989$\pm$13.0998} & \textbf{32.4097$\pm$2.4509} & \textbf{0.9672$\pm$0.0237} \\
\hline
\end{tabular}
\label{tab:generator_comparison}
\end{table}

\subsection{Loss Function Ablation}
Table~\ref{tab:loss_ablation} summarizes the impact of progressively adding each loss component on the MRI-to-CT validation set. Using $\mathcal{L}_{\mathrm{mae}}$ alone results in the lowest performance. Introducing the adversarial term $\mathcal{L}_{\mathrm{adv}}$ yields the largest single-step improvement, reducing MAE by 4.6186 and boosting PSNR by 1.1124dB. Adding $\mathcal{L}_{\mathrm{fm}}$ further improves all metrics, indicating enhanced stability in adversarial training. The perceptual loss $\mathcal{L}_{\mathrm{perc}}$ leads to additional structural fidelity gains, reflected in a +0.3559~dB PSNR increase. Finally, incorporating the segmentation-based masked MAE $\mathcal{L}_{\mathrm{mask}}$ achieves the best overall results (MAE: \textbf{67.4066}, PSNR: \textbf{29.0467}~dB, MS-SSIM: \textbf{0.9245}), demonstrating the benefit of constraining supervision to anatomically relevant regions.

\subsection{Comparison with Baselines}
Table~\ref{tab:generator_comparison} compares GANeXt against convolution-based (3D U-Net~\cite{ronneberger2015unet}, 3D U-Net++~\cite{unetpp++}, MedNeXt~\cite{roy2023mednext}) and transformer-based (SwinUNETR~\cite{hatamizadeh2021swin}) backbones on both MRI-to-CT and CBCT-to-CT synthesis.

On the MRI-to-CT local validation set, GANeXt achieves the best scores (MAE: \textbf{67.4066}, PSNR: \textbf{29.0467}dB, MS-SSIM: \textbf{0.9245}), outperforming the second-best MedNeXt by 2.0457 MAE, 0.8142dB PSNR, and 0.0014 MS-SSIM.On the official challenge validation set, GANeXt$^\dagger$ improves further to a MAE of \textbf{62.6230}, PSNR of \textbf{29.9792}~dB, and MS-SSIM of \textbf{0.9325}, indicating strong generalization to unseen data.

For CBCT-to-CT, the local validation results show GANeXt surpassing all baselines by a large margin (MAE: \textbf{52.1145}, PSNR: \textbf{31.2487dB}, MS-SSIM: \textbf{0.9662}), with its closest competitor, MedNeXt, trailing by a significant 21.7276 in MAE, 3.2622dB in PSNR, and 0.0428 in MS-SSIM. On the official validation set, GANeXt$^\dagger$ achieves its highest performance (MAE: \textbf{48.9989}, PSNR: \textbf{32.4097}~dB, MS-SSIM: \textbf{0.9672}), further underscoring its robustness to domain shift.

\section{Discussion}

The results demonstrate that our GANeXt model, combined with a carefully designed multi-head discriminator, multi-term objective, delivers consistent improvements over both transformer-based and conventional convolutional architectures for sCT synthesis from MRI and CBCT. The strong gains on the official challenge validation sets indicate that GANeXt can generalize effectively to unseen data distributions, which is critical for clinical deployment.

Nevertheless, this study has several limitations. First, we observed that the synthesized CT tends to underestimate intensities in high-HU regions, particularly in dense bony structures. This is likely due to the sparse distribution of high-intensity voxels in the training set, which may limit the model’s ability to learn rare tissue classes accurately. Such bias could compromise downstream applications requiring precise bone depiction, such as dose calculation or surgical planning. Second, as a fully supervised framework, GANeXt’s performance is inherently dependent on the accuracy of deformable registration between the input modality and the reference CT. Even with modern registration algorithms, residual misalignments can introduce spatial and intensity inconsistencies that act as bias during training, potentially limiting generalizability across different scanners, acquisition protocols, or anatomical variations.

\begin{credits}
\subsubsection{\ackname} We thank the SynthRAD2025 challenge organizers for providing the dataset. The authors gratefully acknowledge the scientific support and HPC resources provided by the Erlangen National High Performance Computing Center of the University of Erlangen-Nuremberg. The hardware is funded by the German Research Foundation.

\end{credits}

%
%
\bibliographystyle{splncs04}
\bibliography{mybibliography}
%




\end{document}